\documentclass{article}

    \PassOptionsToPackage{numbers, compress}{natbib}

\usepackage[preprint]{neurips_2024}

\usepackage[utf8]{inputenc} 
\usepackage[T1]{fontenc}    
\usepackage[hidelinks,colorlinks=true,linkcolor=blue,citecolor=blue]{hyperref}    
\usepackage{url}            
\usepackage{booktabs}       
\usepackage{amsfonts}       
\usepackage{amsmath,bm}
\usepackage{nicefrac}       
\usepackage{microtype}      
\usepackage{xcolor}         
\usepackage{adjustbox}
\usepackage{enumitem}
\usepackage{wrapfig}
\usepackage{subcaption}
\usepackage{todonotes}
\usepackage{tcolorbox}

\usepackage{tikz}
\newcommand{\cblock}[3]{
  \hspace{-1.5mm}
  \begin{tikzpicture}
    [
    node/.style={square, minimum size=10mm, thick, line width=0pt},
    ]
    \node[fill={rgb,255:red,#1;green,#2;blue,#3}] () [] {};
  \end{tikzpicture}%
}

\title{LoRA Soups: Merging LoRAs for Practical\\Skill Composition Tasks}

\author{%
  Akshara Prabhakar\thanks{Corresponding author  <\href{mailto:akshblr555@gmail.com}{\texttt{akshblr555@gmail.com}}>}\\
  Department of Computer Science\\
  Princeton University\\
  \And
  Yuanzhi Li \\
  Microsoft Research \\
  \And
  Karthik Narasimhan \\
  Department of Computer Science\\
  Princeton University\\
  \And
  Sham Kakade, Eran Malach \\
  Kempner Institute \\
  Harvard University\\
  \And
  Samy Jelassi\\
  Center of Mathematical Sciences and Applications \\
  Harvard University\\
}

\begin{document}

\maketitle

\begin{abstract}


Low-Rank Adaptation (LoRA) is a popular technique for parameter-efficient fine-tuning of Large Language
Models (LLMs).
We study how different LoRA modules can be merged to achieve \textit{skill composition}---testing the performance of the merged model on a target task that involves combining multiple skills, each skill coming from a single LoRA. This setup is favorable when it is difficult to obtain training data for the target task and when it can be decomposed into multiple skills. First, we identify practically occurring use-cases that can be studied under the realm of skill composition, e.g.\ solving hard math-word problems with code, creating a bot to answer questions on proprietary manuals or about domain-specialized corpora. Our main contribution is to show that concatenation of LoRAs (\textbf{CAT}), which optimally weights LoRAs that were individually trained on different skills, outperforms existing model- and data- merging techniques; for instance on math-word problems, CAT beats these methods by an average of 43\% and 12\% respectively. Thus, this paper advocates model merging as an efficient way to solve compositional tasks and underscores CAT as a simple, compute-friendly and effective procedure. To our knowledge, this is the first work demonstrating the superiority of model merging over data mixing for binary skill composition tasks.\footnote{Code and data are available at \url{https://github.com/aksh555/LoRA-Soups}.}

\end{abstract}

\section{Introduction}
\label{sec:introduction}

Large Language Models (LLMs) have demonstrated impressive capabilities in conversational tasks and general-purpose applications, such as writing emails or answering common questions. However, these general purpose LLMs may have restricted performance on tasks where specific skills and knowledge is required. We primarily focus on skill composition tasks, that necessitate the integration of multiple skills. 

Many industrial applications fit in this framework. 
Consider a company that manufactures ovens and is trying to design a chatbot to answer customer queries about its working and specifics. Directly using a frontier LLM (like \texttt{gpt-4o}) would fail since it lacks knowledge about the company's product. The ideal solution here would be to design an instruction dataset consisting of question-answer pairs about this product and fine-tuning an LLM on it. However, such a data collection and annotation procedure is expensive. Another possible solution is to fine-tune an LLM on a collection of product manuals and then impart it chat abilities by further fine-tuning on an instruction-tuning dataset like Alpaca \cite{alpaca}. We refer to this method as DATA-MIX. Besides this approach being sequential, it suffers from catastrophic forgetting \cite{doi:10.1073/pnas.1611835114}.
Whenever the company creates a new product, they need to redo fine-tuning on this data mixture or create a new question-answer dataset for the former method. 

In this paper, we study \emph{model merging} as an alternative approach. Given a model that is fine-tuned on the manuals and one that possesses question-answering capabilities, we optimally combine their weights to obtain a model that can answer product-specific questions. This approach is more efficient since we merge skill-specific fine-tuned models without any additional data collection or training from scratch. Among the multiple techniques to perform model merging, our framework specifically builds on LoRA \cite{hu2021lora} (a review of the LoRA method is in \autoref{app:lora}), a fine-tuning technique that consists of adding a low-rank update to a few layers in the model. In this context, model merging consists of combining the LoRA weights from different models. 

Prior works \cite{gu2024mix,shah2023ziplora,yang2024lora} have investigated LoRA merging in computer vision where each skill is a visual concept or style and the objective is image generation. On the other hand, natural language tasks are more challenging since identifying the skills needed for solving a task is not always clear. On natural language tasks, most prior works \cite{buehler2024x,feng2024mixture,luo2024moelora,muqeeth2024learning,wu2023mole} merged LoRAs with the objective of multitask learning. In this setting, the individual LoRA modules are trained on (potentially) independent tasks and the merged model is tested on the original tasks. A successful model retains the skills of each individual LoRA. Differently, \citet{huang2023lorahub} devise LoraHub, a strategy to merge LoRAs for cross-task learning. By finetuning LoRAs on FLAN \cite{longpre2023flan}, they achieve performance equivalent to few-shot prompting on some Big-Bench Hard (BBH) tasks \cite{suzgun2022challenging}. Closer to our work,  \citet{akiba2024evolutionary} merge specialized LMs in Japanese and in math \cite{cobbe2021training} to solve word-math problems in Japanese \cite{shi2022language}. 
Though this can be viewed as a skill composition task, they focus on studying only one such task, and their contribution is an evolutionary algorithm for merging models. 

\textit{Given LoRAs trained on specialized domains (biology, math, code, reading comprehension, question-answering), is it possible to merge them to effectively solve a new problem that requires a combination of these domains?} 

We underline that most settings we consider are \emph{out-of-domain} since the specialized LoRAs have been trained on datasets that are very different from the target task. To our knowledge, this is the first paper exhibiting model merging is superior to data mixing for binary skill composition problems. Our key contributions are summarized as follows:

\begin{figure}
\vspace*{-0cm}

\begin{subfigure}{.65\textwidth}
\centering
    \hspace*{-.5cm}\includegraphics[width=.95\linewidth]{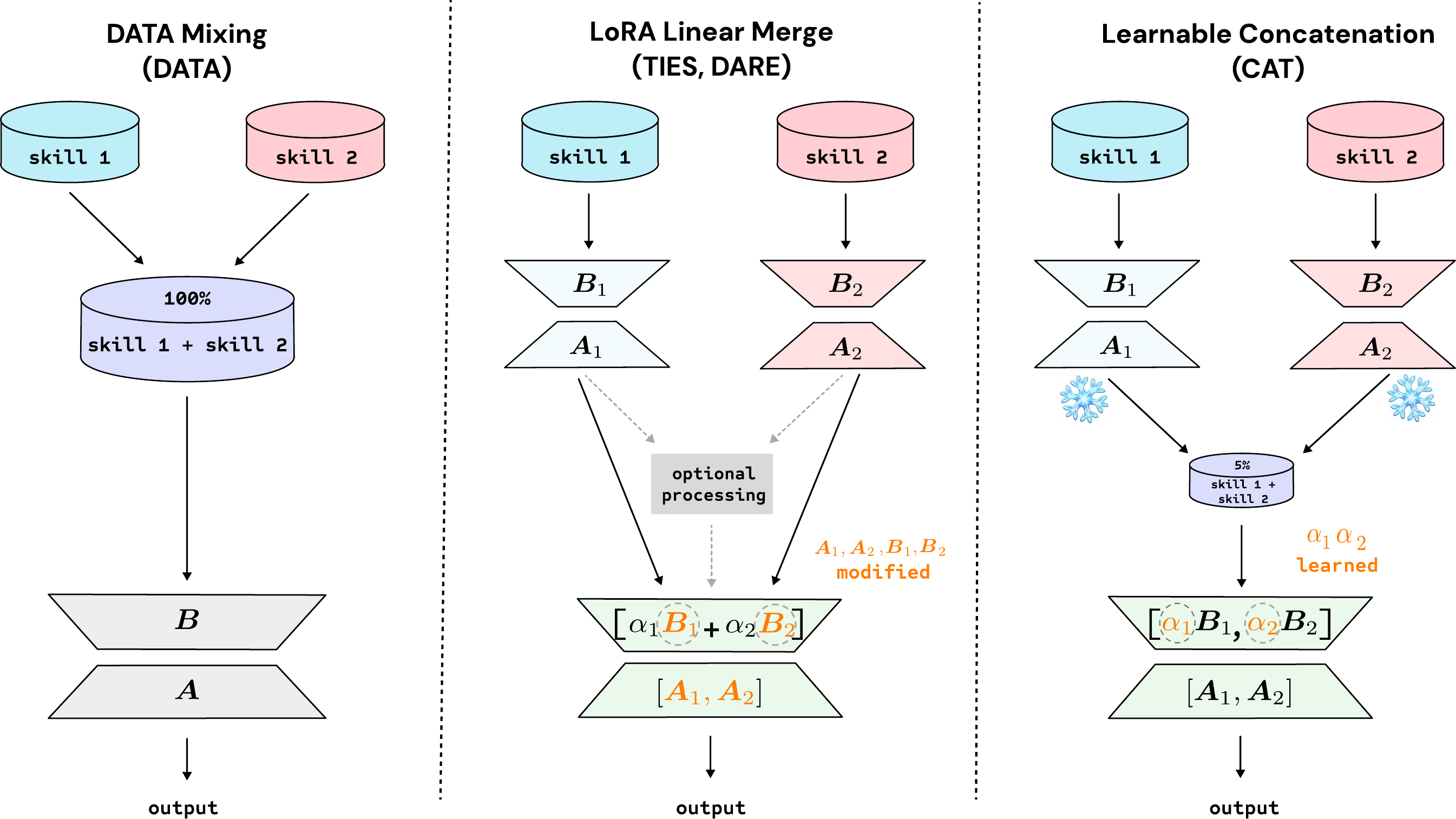}
    \caption{\small Comparison of CAT with DATA-MIX and Linear methods.}
    \label{fig:methods_comparison}
    \end{subfigure}
\begin{subfigure}{0.32\textwidth}
\centering
    \hspace*{-.5cm}\includegraphics[width=1.3\linewidth]{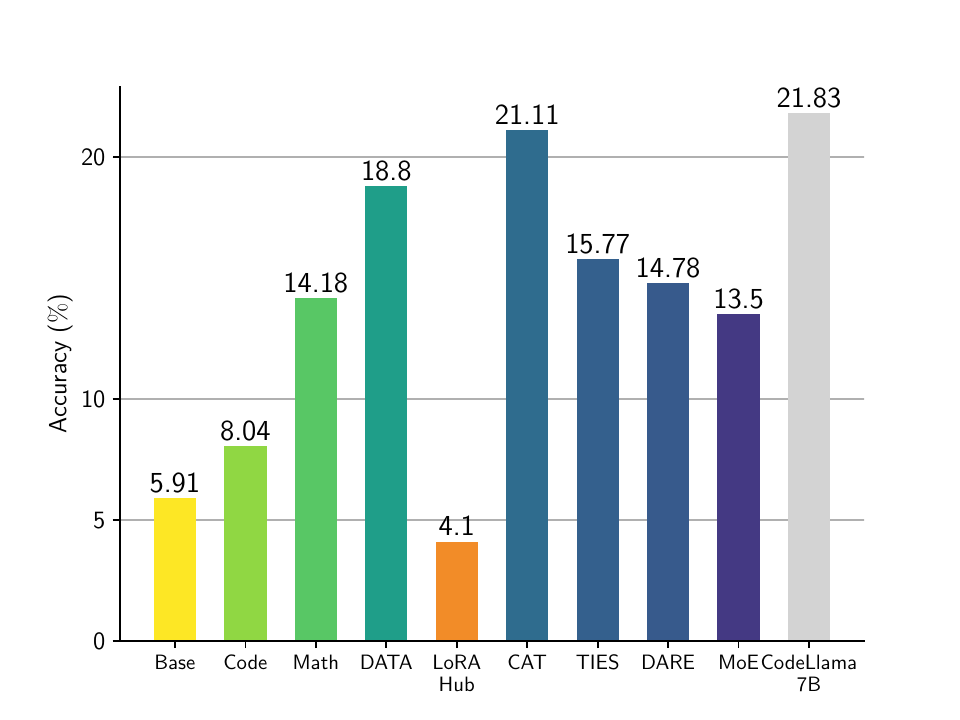}
    \caption{\small Performance of different methods on solving hard math-word problems with code.}
    \label{fig:math_code_scores}
\end{subfigure}
\caption{Method and performance overview of Learnable Concatenation (CAT).}\label{fig:preview}
\end{figure}


\begin{itemize}[leftmargin=*, itemsep=1pt, topsep=1pt, parsep=1pt]
    \item[--] In \autoref{sec:setting}, we define the \emph{skill composition} problem and show some practical applications in domains spanning code, science, robustness, and in-house use cases. 
    \item[--] We introduce Learnable Concatenation (\textbf{CAT}), a LoRA merging technique that involves a simple weighted average of the encompassed skill LoRAs. While concatenating LoRAs as a technique already exists, our contribution lies in cheaply learning the weights layer-wise and applying this to compositional natural language tasks.
    \item[--] In \autoref{sec:experiments}, we perform a comprehensive evaluation of several baselines and demonstrate that \textbf{CAT} achieves better binary skill composition than both existing merging methods and data mixing. 
    \item[--] In \autoref{sec:limitations}, we discuss some limitations of our approach. Our main limitation is that we focus on skill composition problems that require 2 skills and an interesting direction would be to extend this to more skills. 
\end{itemize}

\section{Related Work}
\label{sec:related}
\textbf{Merging methods. } The traditional approach to learning multiple skills/tasks simultaneously is joint training on a mixture of task datasets \cite{caruana1997multitask}. Over the years, several works have improved this multi-task learning approach \cite{misra2016cross,sun2020adashare,sener2018multi,guo2020learning,ma2018modeling,chen2018gradnorm,sener2018multi}. As data collection for specialized tasks, and training large models from scratch get more expensive; coupled with the rapid expansion in the availability of well-trained open-source models -- model merging has emerged as a convenient way of building powerful models from existing ones \cite{goddard2024arcees,mergoo2024}. 
The richly studied simplest way of merging by averaging model weights \cite{utans1996weight,smith2017investigation,garipov2018loss,izmailov2018averaging} paved the way to linear weight averaging \cite{wortsman2022model}. Some methods like  Fisher Merging \cite{matena2022merging} and RegMean \cite{jin2023dataless} need training data based pre-computations to measure individual parameter importance but these are highly memory and data intensive.
Expanding on weight averaging, Task Arithmetic \cite{ilharco2022editing} involving the creation and combination of task vectors facilitated multi-task learning. While this weight interpolation was heavily used for merging image generation models, 
recent methods like TIES \cite{yadav2024ties} and DARE \cite{akiba2024evolutionary} reset redundant parameters, resolve sign conflicts,
and exclusively merge parameters that exhibit sign-consistency, and SLERP \cite{white2017sampling} by spherical linear interpolation build upon this for language models. In all these methods, the coefficients governing the model merging are determined by trial-error; while works in the vision domain demonstrate the pivotal role played by these coefficients \cite{yang2023adamerging}. In contrast to the cumbersome grid-searching, our method CAT learns these coefficients layer-wise with access to very few examples in the natural language domain.

\textbf{LoRA merging methods. }
PEM composition \cite{zhang2023composing} adopts the task arithmetic framework to incorporate the merging of LoRAs. Recently, the vision community witnessed the widespread application LoRAs   \cite{buehler2024x,feng2024mixture,luo2024moelora,muqeeth2024learning,wu2023mole,zhong2024multilora,yang2024loracomposer} as an effective approach to multi-task learning and composing styles and subjects \cite{shah2023ziplora}. Many of these utilize Mixture of Experts (MoE) \cite{buehler2024x,feng2024mixture,luo2024moelora,wu2023mole} based architectures having input-dependent learnable routers. These models have been primarily used in the context of multitask learning. Apart from multi-task learning  \cite{ilharco2022editing,li2022branch,yadav2023tiesmerging,wu2023mole,feng2024mixture,yang2023adamerging}, merging models have served various use cases -- ensembling to improve generalization on the target task \cite{izmailov2018averaging,arpit2022ensemble,gupta2020stochastic}, federated learning \cite{mcmahan2017communication}, model compression \cite{li2023merge}. In this work, we study LoRA model merging for \textit{skill composition} tasks.





\section{Setting}
\label{sec:setting}


\subsection{Skill composition}\label{sec:problems}


Most of the downstream tasks used to evaluate LLMs require mastering multiple skills to be solved. Skill here refers to specific capabilities that the LLM needs for customization to downstream use cases. These skills can be acquired from knowledge source like textbooks and manuals or from foundational datasets designed for arithmetic, coding, etc. For instance, achieving high score in the GSM8k benchmark \cite{cobbe2021training} requires good commonsense reasoning and arithmetic skills. In this paper, we focus on downstream tasks where  composition can be ensured and isolated, and we mainly focus on tasks that require two skills. Skill composition is challenging because, not only does the model need to ``know'' the different skills, but it also needs to understand the appropriate context for applying each skill. We now present some skill composition examples of practical interest and \autoref{table:task_summary} summarizes them.

\begin{wrapfigure}[10]{r}{0.45\textwidth}
\vspace{-1cm}
\includegraphics[width=0.5\textwidth]{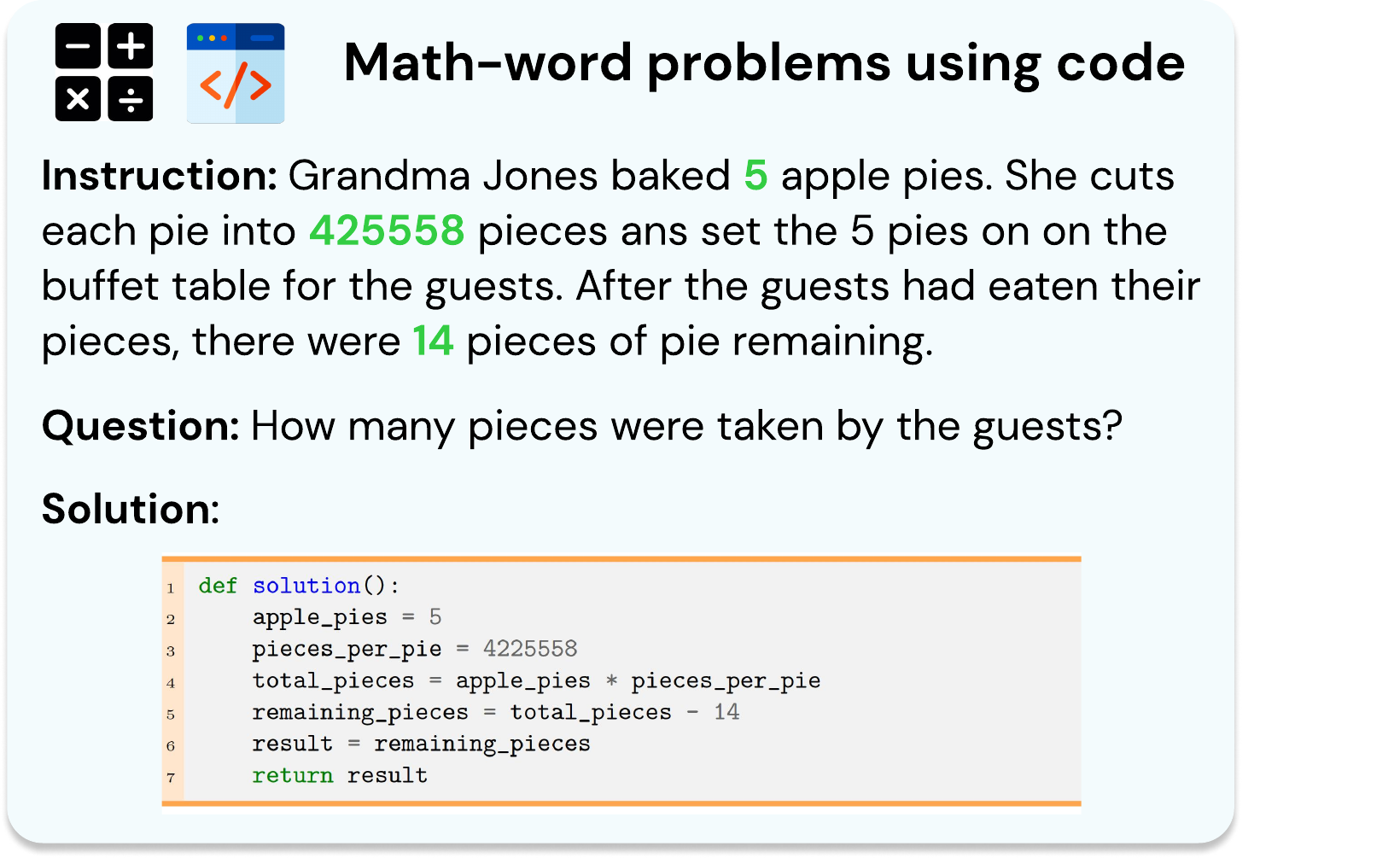}
\caption{\small An example of solving hard word-math problem using code.}\label{fig:math_code_data}
\end{wrapfigure}

\paragraph{Hard math-word problems.} Prior works noticed that when fine-tuning a large language model on GSM-8k \cite{cobbe2021training}, the resulting model performs poorly on GSM-Hard \cite{gao2023pal}, a dataset that gathers similar problems but with more complex arithmetic operations. The program-aided approach by Gao et al.\  \cite{gao2023pal} is one solution to address this issue: the LLM reads the GSM problem and generates programs as the intermediate
reasoning steps, but offloads the solution step to a Python interpreter. In this case, the model must excel in both mathematical skills to first reason about the problem and also coding skills to translate its reasoning to code. Therefore, in order to improve the accuracy on GSM-hard, we set the first skill to be MetaMathQA \cite{yu2023metamath} -- an augmented version of the GSM-8k dataset and the second skill to be Code-Alpaca \cite{codealpaca}. 


\begin{wrapfigure}[10]{r}{0.45\textwidth}
\vspace{-.5cm}
\includegraphics[width=0.5\textwidth]{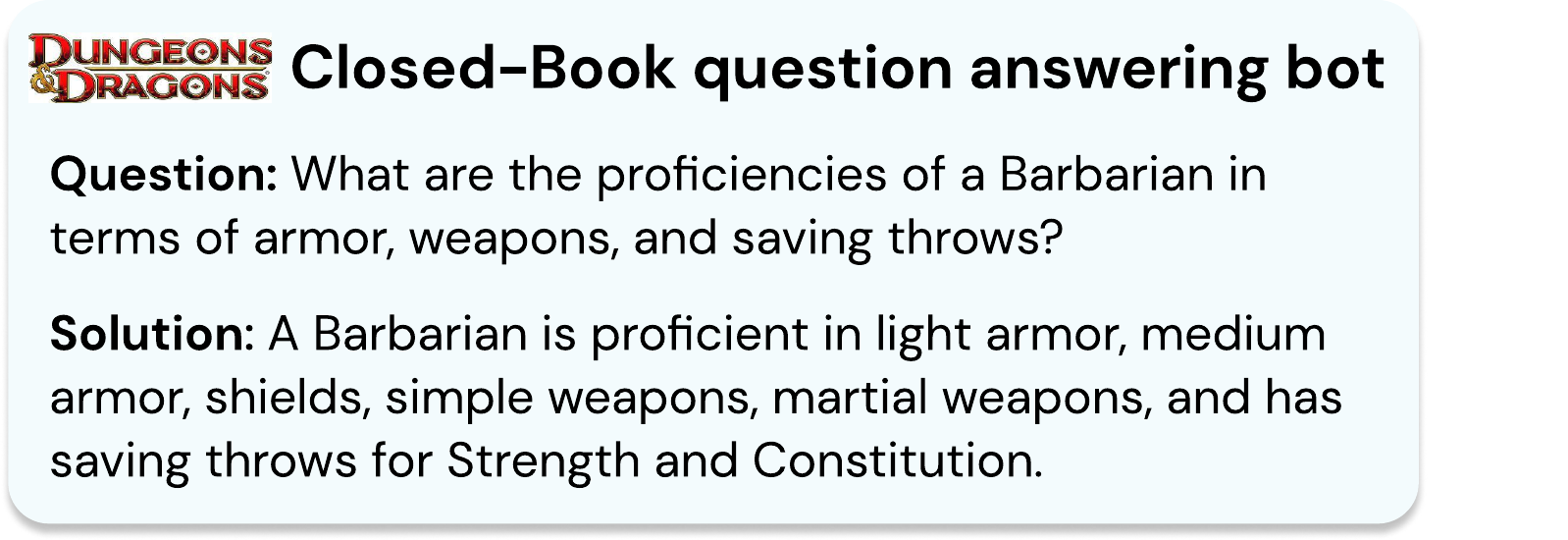}
\caption{\small An example of question-answering task based on \textit{Dungeons \& Dragons} game manual.}\label{fig:qabot_examples}
\end{wrapfigure}

\paragraph{QABot on proprietary manuals.}  Chat models are language models that aim to facilitate conversational interactions. Similarly to LLMs, they are first pre-trained on large language datasets. Then, they are finetuned on instruction-tuning datasets  \cite{vicuna2023,conover2023free,kopf2024openassistant,alpaca,xu2023baize,xu2023wizardlm,zhou2024lima}. While general-purpose chatbots are useful, an institution or corporation may desire to have a \emph{specialized} question-answering bot/assistant that addresses domain-specific questions. Examples of this case may be a university that wants a bot that answers questions about quantum physics or a refrigerator retailer that wants a bot to answer questions about their last device. The aforementioned approach may fall short since when fine-tuning a chat model on a specialized document, it may lose its conversational abilities. Another solution would be to obtain an instruction-tuning dataset that covers the specialized material but this may be too costly. On the other hand, model merging seems to be a convenient solution to address this problem: we train one LoRA model on a general instruction-tuning dataset and another one on the specialized document. Therefore, we set the first skill to be the Alpaca dataset \cite{alpaca} and study two cases where the second skill is: (1) a molecular biology university textbook \cite{bergtrom2022basic} from \url{https://bio.libretexts.org/} and (2) a manual of the "Dungeons and Dragons" game \cite{gygax1974dungeons}. 


\begin{wrapfigure}[14]{r}{0.45\textwidth}
\vspace{-.5cm}
\includegraphics[width=0.5\textwidth]{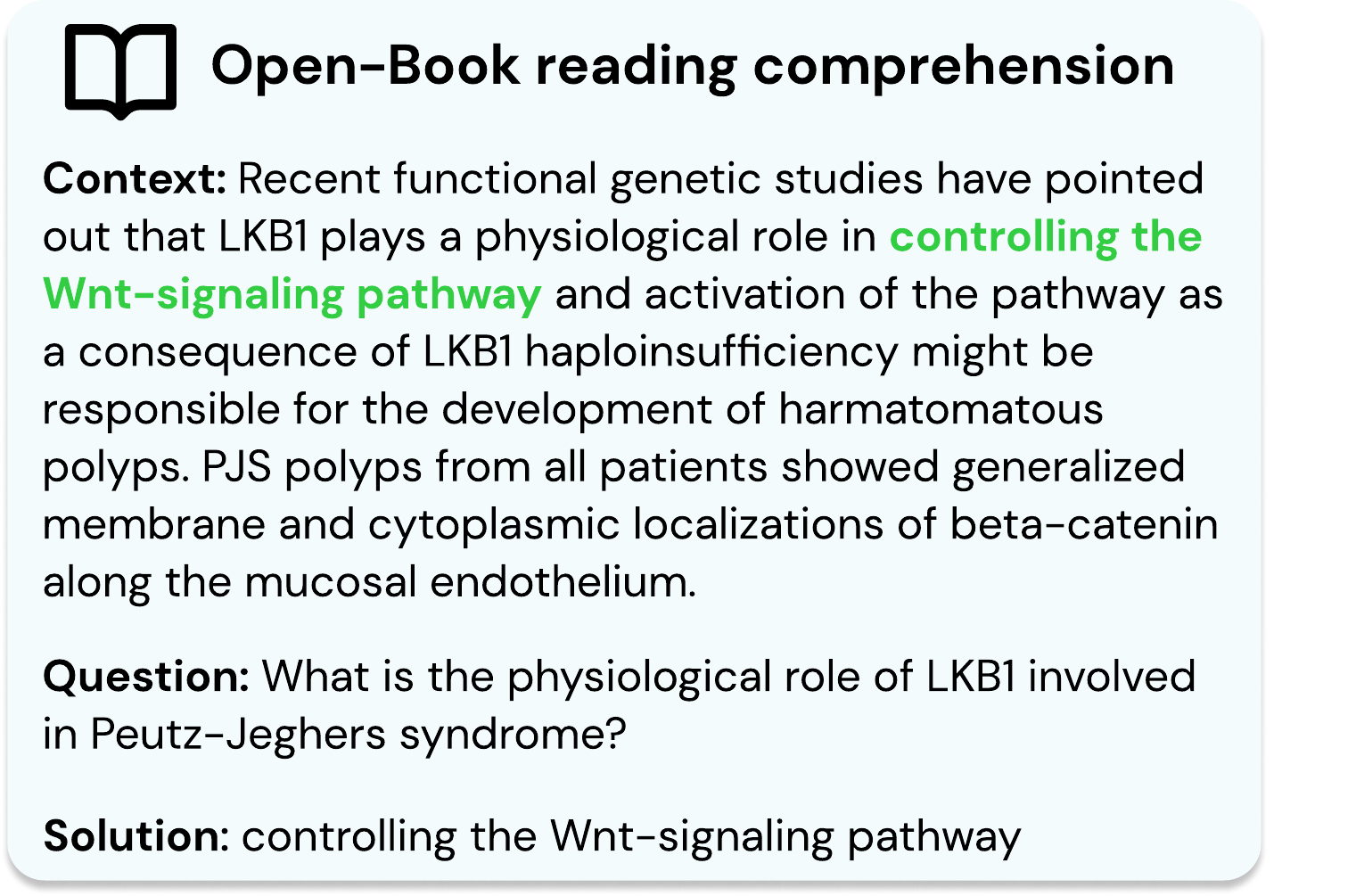}
\caption{\small An example of the reading comprehension task in the biomedical domain.}\label{fig:reading_comprehension_ata}
\end{wrapfigure}

\paragraph{Reading comprehension on technical documents.} Medical reports and law contracts are very challenging to read for non-specialist users: they are usually very long and involve a myriad of technical jargon. Having a language model that can read these documents and answer questions would be very useful. For this reason, we train one LoRA model on SQuAD \cite{rajpurkar2018know} to acquire the reading comprehension skill and another LoRA on open-source biology textbooks \cite{grewal2023human,klymkowskybiofundamentals,nelson2021medical,singh2014genetics} collected from \url{https://libretexts.org}. We test our models on the BioASQ-QA dataset \cite{krithara2023bioasq}, a reading comprehension dataset on biomedical engineering. 

\subsection{Merging methods}\label{sec:merging}

\autoref{fig:methods_comparison} details the various merging methods in the literature and the proposed CAT method.


\paragraph{Data mixing (DATA-MIX).} The naive approach for solving these tasks is to train on a mixture of datasets containing different skills. This method does not merge models but merges datasets. We fine-tune a \emph{single} model with LoRA weights $(A,B)$ where $A,B\in\mathbb{R}^{d\times kr}$ on the concatenation of datasets and thus simultaneously teach the $k$ skills to the model. Data mixing is a very expensive approach in practice. Indeed, whenever we want to add a new skill to the model, we need to retrain the model on an updated mixture of datasets. Model merging offers a solution to this problem. We train one model per skill and then merge all of them to solve the compositional task.
We set the rank to $kr$ (and not $r$) in order to ensure a fair comparison with the other merging methods.


Next, we review the different merging schemes and report their skill composition performance in the next sections. We distinguish two main classes of LoRA merging techniques: the concatenation of LoRAs and the linear merging of LoRAs. These two methods differ in whether we linearly combine the parameters or the LoRA updates.

\paragraph{Concatenation of LoRAs.} This method consists of a linear combination of the LoRA updates i.e.\ $\Delta W^l = \alpha_1^l B_1A_1^{\top}+\alpha_2^l B_2A_2^{\top}$, where $\alpha_1^l,\alpha_2^l\in[0,1]$ are merging coefficients for layer $l$. We refer to this method as concatenation of LoRAs. In what follows, we study three variants of this method that differ by their merging coefficients definition: 
\begin{itemize}
    \item[--] \textbf{Learnable concatenation (CAT)} (introduced in this paper): in this variant, we set $\alpha_1^l,\alpha_2^l$ as trainable parameters. This distinguishes us from other methods like TIES, DARE, LoRA Hub that learn static values for every layer $l$ of the network. Once the LoRA modules are separately trained, we add a step where we only train the merging coefficients on a small mixture of the datasets.
    \item[--] \textbf{Mixture of Experts (MoE)} \cite{buehler2024x,feng2024mixture,luo2024moelora,muqeeth2024learning,wu2023mole}: the main difference of this method compared to the previous ones is that the merging coefficients are \emph{input-dependent}. Indeed, we have a trainable router parameter $W_r^l\in \mathbb{R}^{d\times 2}$ which computes the logits $h^l(x)=W_r^{l\top} x$. Then, the merging coefficients are defined as:
    \begin{equation*}
    \begin{aligned}
       \alpha_1^l = \frac{e^{h^l(x)_1}}{e^{h^l(x)_1}+e^{h^l(x)_2}}, \alpha_2^l = \frac{e^{h^l(x)_2}}{e^{h^l(x)_1}+e^{h^l(x)_2}}
    \end{aligned}
    \end{equation*}
\end{itemize}

\paragraph{Linear merging of LoRAs.} This method consists of a linear combination of the LoRA parameters i.e. $\Delta W^l = (\alpha_1^l B_1+\alpha_2^l B_2)(\alpha_1^l A_1+\alpha_2^l A_2)^{\top}$. \textbf{We note that all these methods use the same static weights $\bm{(\alpha_1, \alpha_2)}$ for every layer.} Compared to the concatenation of LoRAs, this method involves additional cross-terms.
In what follows, we study three variants of this method that apply a series of preprocessing steps on the LoRA parameters before applying the update. 
\begin{itemize}
    \item[--] \textbf{TIES} \cite{yadav2024ties}: we prune the smallest values of $(A_k,B_k)$ for $k\in\{1,2\}$ and retain the top values based on the specified density fraction $\lambda\in[0,1].$ Then, we calculate the majority sign mask from the pruned LoRA weights by summing all their parameter values and storing the sign of this sum. Lastly, the LoRA weights are multiplied by weights $(\alpha_1,\alpha_2)$. Finally, we apply the linear merging update based on the stored majority sign. Here, $(\lambda,\alpha_1,\alpha_2)$ are hyper-parameters.
    \item[--] \textbf{DARE} \cite{yu2023language}: we first randomly prune the values of the LoRA parameters $(A_k,B_k)$ for $k\in\{1,2\}$ based on a density $\lambda\in[0,1]$. Then, we rescale the pruned LoRA weights by $1/\lambda$. Finally, we apply the linear merging update. Here, $(\lambda,\alpha_1,\alpha_2)$ are hyper-parameters.
    \item[--] \textbf{LoRA Hub} \cite{huang2023lorahub}: this method was primarily proposed to select and assign weights to the constituent LoRA modules that would help solve an unseen test task. Here $(\alpha_1,\alpha_2)$ are learned from a few (5) examples from the target task in a gradient-free manner.

\end{itemize}
Most of these aforementioned methods have been introduced in the literature for solving the multitask problem, i.e.\ to perform well simultaneously on $k$ skills, when tested independently. In this paper, we study how these methods perform in \emph{skill composition}.

\section{Experimental details}
\label{s:exp_details}


\paragraph{Skill Finetuning.} Our base model is a Llama-7b \cite{touvron2023llama} and set the LoRA rank $r=32$, LoRA alpha $ = 64$, LoRA dropout $ = 0.05$ and the target modules to be $\texttt{\{"q\_proj", "v\_proj", "k\_proj", "up\_proj","down\_proj"\}}$. 
We finetune the individual LoRAs using AdamW \cite{loshchilov2017decoupled} for $3$ epochs, with a learning rate $3\mathrm{e}{-4}$, $100$ warmup steps, a linear decaying schedule, batch size in $\{4,8\}$ and gradient accumulation $4$. We set the precision to $\texttt{float16}$. Our loss is the standard next-token prediction loss. When finetuning on the instruction datasets, we mask the instruction so that we only penalize wrong token predictions at the level of answers. When finetuning on textbooks, we do not apply any masking.
We followed the implementation of \textit{mergoo} \cite{mergoo2024} for MoE which is based on recent MoE for LoRAs \cite{feng2024mixture,buehler2024xlora}. 

\paragraph{DATA-MIX.} 
For DATA-MIX, we train LoRA modules with a rank $2r=64$, to ensure an equal number of trainable parameters with the merging methods. For LoRA Hub, we used at least 5 few-shot examples for learning the weights. Training a model on a mixture of datasets in \autoref{sec:reading_comprehension}, \autoref{sec:chatbot} is not straightforward, as different examples have different masking schemes, which makes standard data mixing fail. When training with alternating between one batch of each dataset type we observed poor performance. Hence, we perform continual training by fine-tuning for $3$ epochs on the chapter/manual, merge the weights with the pre-trained model weights, followed by $1$ epoch of fine-tuning on the instruction-following dataset (SQuAD/Alpaca) similar to \cite{wu2024pmc}.

\paragraph{CAT.} 
We freeze the trained LoRA skill modules and train $\alpha_1^l,\alpha_2^l$ on a dataset made by selecting the minimum of $5\%$ of the data points from both skill 1 and skill 2. This additional step only runs for $1$ epoch with a learning rate of $1\mathrm{e}{-4}$. The best model is chosen as the one that does best on the validation set.

\paragraph{Hyperparameters.} The $(\lambda, \alpha_1, \alpha_2)$ hyperparameter values for TIES, DARE are chosen by doing a sweep over $\lambda \in [0,1], \alpha_1 \in [1,2], \alpha_2 \in [1,2]$ in increments of $0.2$ and we report the best results. 
At inference time, we generate answers in an autoregressive fashion setting \texttt{temperature} to $0.01$, \texttt{max\_new\_tokens} to $200$, with nucleus sampling probability \texttt{top\_p} as $0.95$. 

\autoref{table:task_summary} summarizes our skill composition tasks, dataset sources, and evaluation criterion. Further training details and compute used are presented in \autoref{appx:compute}. 

\begin{table}[t]
\centering
\resizebox{\textwidth}{!}{%
\begin{tabular}{lllll}
\toprule
    \bfseries Task
    & \bfseries Skills Composed
    & \bfseries Test Dataset 
    & \bfseries Evaluation Metric \\
\midrule
    Hard math-word & \{math, code\} & GSM-Hard \cite{gao2023pal} (1319) & Exec. Accuracy \\
    Question-answering & \{biology/game, closed-book QA\} & Open-source book/manual \S\ref{appx:qa_data} (95) & Accuracy (GPT-4) \\
    Reading comprehension & \{biomedical, open-book QA\} & BioASQ-QA \cite{krithara2023bioasq} (200) & Elo Rating (GPT-4)\\
    Prompt robustness & \{$prompt_i$, $prompt_j$\} & SuperNaturalInstructions \cite{supernaturalinstructions} (417) & Accuracy \\
\bottomrule
\end{tabular}}
\vspace{0.5em}
\caption{Rundown of our skill-composition tasks, source of test datasets, and evaluation criterion.}
\label{table:task_summary}
\vspace{-1.5em}
\end{table}
\section{Experiments}\label{sec:experiments}

\subsection{Super-linear improvement when solving GSM-Hard with code}\label{sec:math_code}


\begin{wrapfigure}[20]{r}{0.45\textwidth}
\vspace{-.6cm}
\includegraphics[width=0.45\textwidth]{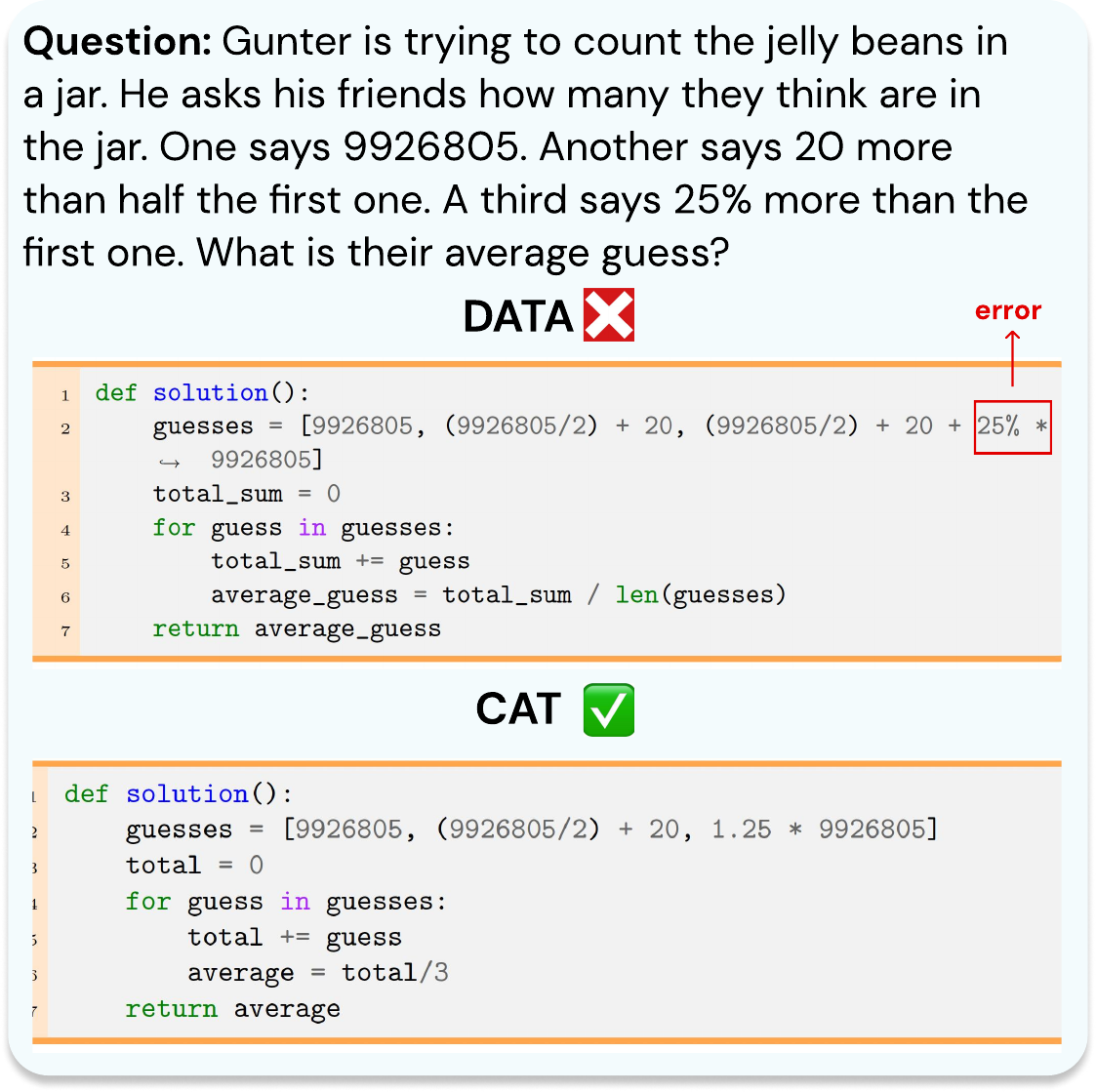}
\caption{\small CAT vs. DATA solving a GSM-Hard problem. DATA makes frequent coding errors. }\label{fig:math_code_exampless}
\end{wrapfigure}

\paragraph{Evaluation setup.} 
We evaluate the ability to solve hard math-word problems in GSM-Hard using code. The model takes in a math word problem, mathematically reasons, and then outputs a corresponding Python function whose return value is the answer to the problem. We run this solution in the Python interpreter and use it to test. This setup is motivated by the fact that while finetuning a single LoRA on MetaMathQA yields 55\% accuracy on the standard GSM-8k, the performance on GSM-Hard is much lower at 14.18\%. Though GSM-Hard and GSM-8k problems share the same structure, they differ in the more complex arithmetic operations involved in GSM-Hard, due to which access to a calculator-like tool aids performance. 

\paragraph{Baselines.}  Base (Llama-7b with 8-shot PAL); Skill LoRAs: Math (trained on MetaMath) with 8-shot PAL, Code (trained on Code Alpaca). DATA-MIX (trained on [MetaMath; Code Alpaca]); LoRA Merging: TIES, DARE, MoE.

\paragraph{Results.} \autoref{fig:math_code_scores} illustrates that finetuning on the concatenation of MetaMathQA and Code-Alpaca -- which corresponds to the DATA strategy is effective since the accuracy increases by 32\% over Math (from $14.18\% \rightarrow 18.8\%$) and the model effectively exploits the synergies between natural and programming language \cite{xu2023lemur}. However, we note that DATA does not always conform to outputting code despite 8-shot PAL prompting, while model merging methods do so $\geq 99\%$ times.

Regarding model merging, Yadav et al.\ \cite{yadav2024ties} and Yu et al.\ \cite{yu2023language} report that TIES and DARE are worse than DATA in the multitask setting. We observe that this finding also holds in the skill composition of math and code. Lastly, CAT is the best method on GSM-Hard; \autoref{fig:math_code_exampless} shows a qualitative example comparing CAT to the next best method DATA. Additionally, despite being fine-tuned on a relatively much smaller general code dataset i.e. Code Alpaca, it matches the performance of Python specialized Code Llama - Python 7b \cite{roziere2023code}, which highlights the effectiveness of linearly combining LoRA updates.

\begin{wraptable}[6]{r}{0.45\textwidth}

\vspace{-.3cm}

\resizebox{.46\columnwidth}{!}{%
\begin{tabular}{l*{4}{c}}
\toprule
 \bfseries   & \bfseries No Code& \bfseries Code  \\
\midrule
\bfseries No Math     & 5.91\%               & 8.04\% \bm{\textcolor{gray}{(+36\%)}}          \\
\bfseries Math   & 14.18\%    \bm{\textcolor{gray}{(+140\%)}}            & 21.11\%   \bm{\textcolor{gray}{(+257\%)}}      \\
 \bottomrule
\end{tabular}
}
\caption{\small Super-linear improvement with CAT.}\label{tab:super_linear_improvement}
\vspace{-1.5em}
\end{wraptable}

\paragraph{Super-linear improvement.} \autoref{tab:super_linear_improvement} demonstrates that with CAT, we achieve a super-linear improvement on GSM-Hard. The performance improvement when finetuning on both math and code with respect to the base model (257\%) is superior to the sum of the improvements on code only (36\%) and math only (140\%). We note that such super-linear improvement essentially proves that the merged model is able to leverage the capabilities of the individual models when solving some of the problems. Indeed, if  Math solves 8\% more problems than the base model, and Code solves 2\% more problems, then the total number of problems solved by both models is at most 10\% more than the original model, with a total of 16\% solved problems\footnote{We make the reasonable assumption that the finetuned models are solving new problems, but not degrading the performance on the problems already solved by the base model.}. The merged model, however, solves 21\% of the problems, meaning that at least 5\% of the problems it solves are not solved by either of the individual models. The solution to these problems must therefore arise from combining the knowledge of both models.

\subsection{Building specialized question-answering bots (QABots)}\label{sec:chatbot}

\paragraph{Evaluation setup.} We test the closed-book question-answering capability (i.e. no access to the book/manual about which questions are asked) by considering two settings: (1) simple -- university biology textbook chapter, (2) hard -- nuanced \textit{Dungeons \& Dragons} game manual. Details on the preparation of the dataset and judging are discussed in \autoref{appx:qa_data}, \autoref{appx:chatbot}. In contrast to \autoref{sec:reading_comprehension}, this is closed-book QA as we have access to the context during inference.

\paragraph{Baselines.} Skill LoRAs: Biology/Manual (trained on textbook/manual), Alpaca (trained on Alpaca). Apart from the usual LoRA merging baselines, we include two additional relevant ones: retrieval using \texttt{langchain} based \texttt{RetrievalQAChain} retrieval using (1) \texttt{Llama-7b} and (2) \texttt{Llama2-7b-chat} models. Here the documents are embedded using sentence transformer \texttt{sentence-transformers/all-MiniLM-L6-v2} and stored in a FAISS vector store. These are indeed open-book but are included to benchmark merging methods and get an upper bound.

\begin{wrapfigure}[12]{r}{0.5\textwidth}
\vspace{-1.3cm}
\includegraphics[width=0.5\textwidth]{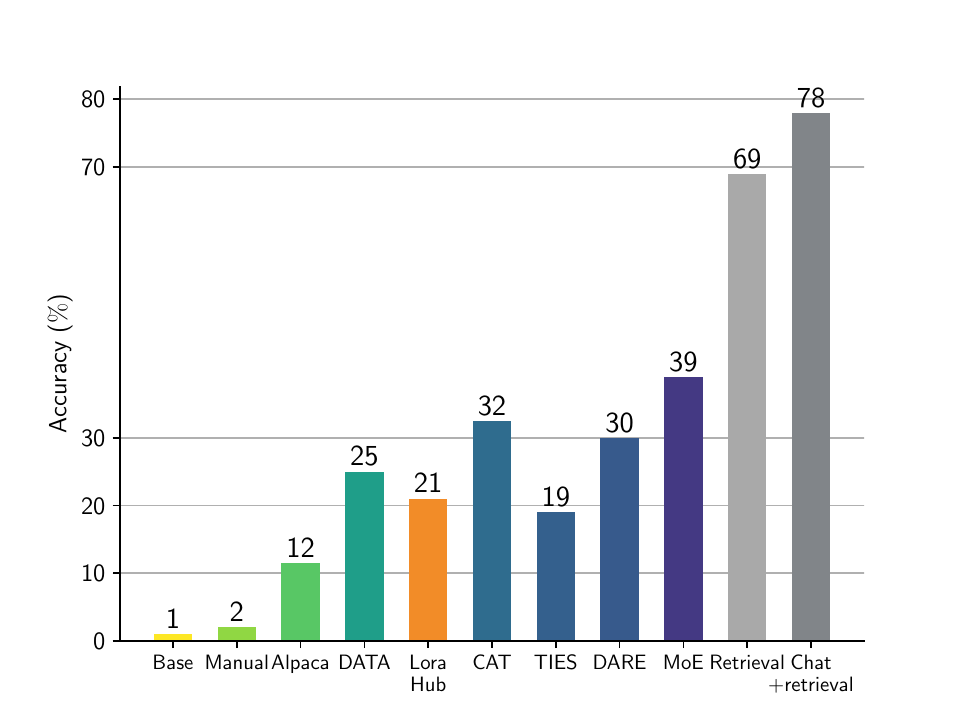}
\vspace{-2em}
\caption{\small Performance on closed-book game QA.}
\label{fig:chatbot_games}
\end{wrapfigure}

\paragraph{Results.} 
\autoref{fig:chatbot_games} shows the performance on the harder game setting. We observe that CAT beats most merging and data mixing but we note the scope for improvement compared to retrieval methods. Here as well, we note the greater need for having the instruction-following skill than the game knowledge due to Alpaca's superior performance; as well as from Llama-chat beating base Llama. Again this highlights how composing the two skills boosts performance. The results on the simpler biology chapter setting are provided in (\autoref{appx:chatbot}, \autoref{fig:chatbot_bio}). 


\subsection{Reading comprehension on technical documents}\label{sec:reading_comprehension}

\paragraph{Evaluation setup.} We test the open-book question-answering ability (i.e. context is accessible to the model) on BioASQ by choosing the ``factoid'' subset of questions to align with the format of questions seen in SQuAD and restricting to examples whose ``Context'' field is less than 500 tokens. Indeed, we observed that the model suffers from length generalization: when the context length is much larger than the one seen during training, the model does not output any answer.
Since the ground truth answers in BioASQ are very long and sometimes contain more details than what is provided in the context, we observe low results in exact matching and F1 scores (see \autoref{appx:reading_comp} \autoref{tab:bioasq_f1}). To alleviate this issue, we use GPT-4 as a judge \cite{zheng2024judging}: given the answers generated by a pair of models, we ask it to score the two in terms of relative correctness to the gold reference answer. This gives a table that records all the pairs of methods, the number of ties, and wins by each model. We report the pairwise win fractions (the ratio of a model's wins over 200 questions) and the ELO rating \cite{elo1978rating} (see \autoref{appx:reading_comp} for prompts and more details). 

\begin{figure}
\vspace*{-0cm}

\begin{subfigure}{.32\textwidth}
    \hspace*{-.5cm}\includegraphics[width=.95\linewidth]{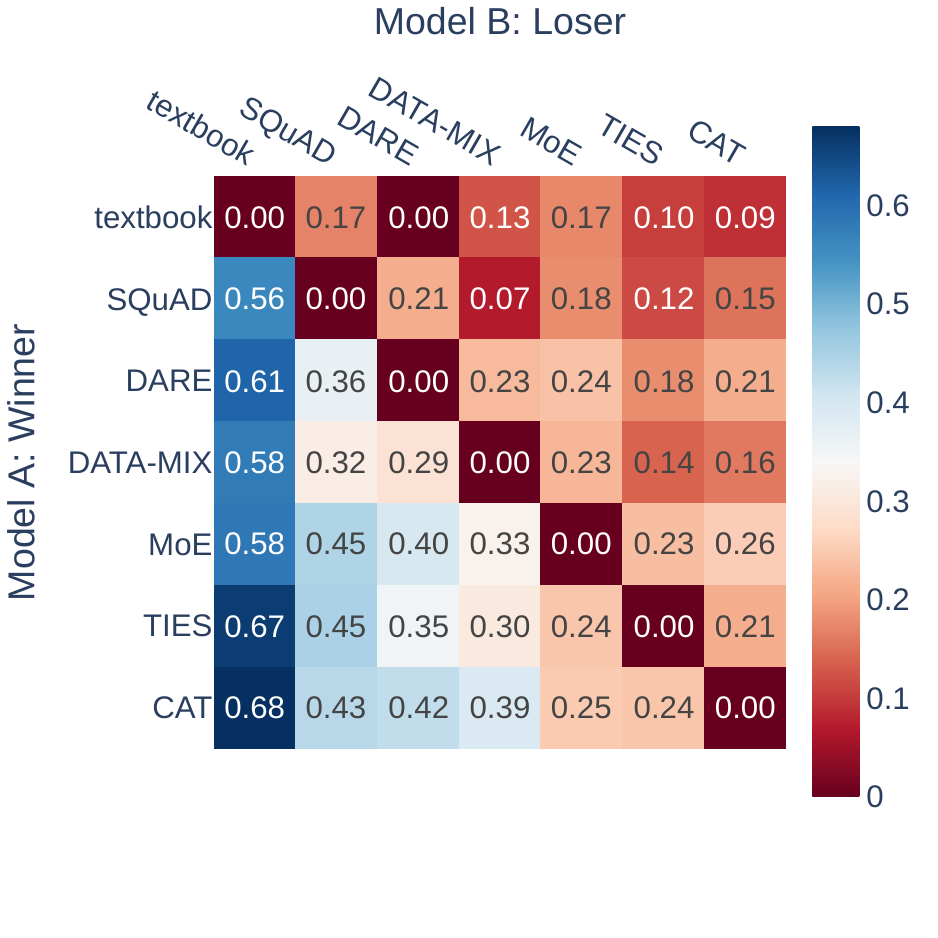}
    \caption{\small Pairwise win ratio of \\Model A w.r.t. Model B.}
    \label{fig:pairwise_win_ratio}
    \end{subfigure}
\begin{subfigure}{0.32\textwidth}
\captionsetup{singlelinecheck=false} 
    \hspace*{-.5cm}\includegraphics[width=1.2\linewidth]{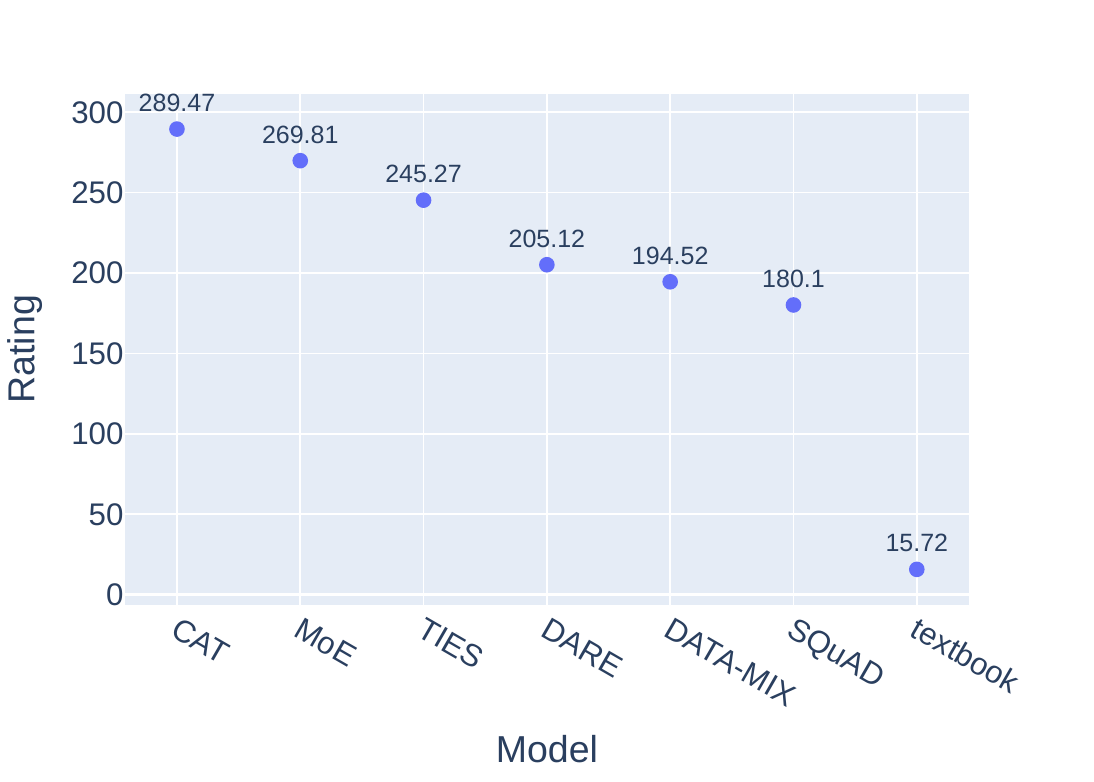}
    \caption{\small ELO Ratings of models\\ on reading comprehension.}

    \label{fig:elo_ratings}
\end{subfigure}
\begin{subfigure}{0.32\textwidth}
    \hspace*{.5cm}\includegraphics[width=1\linewidth]{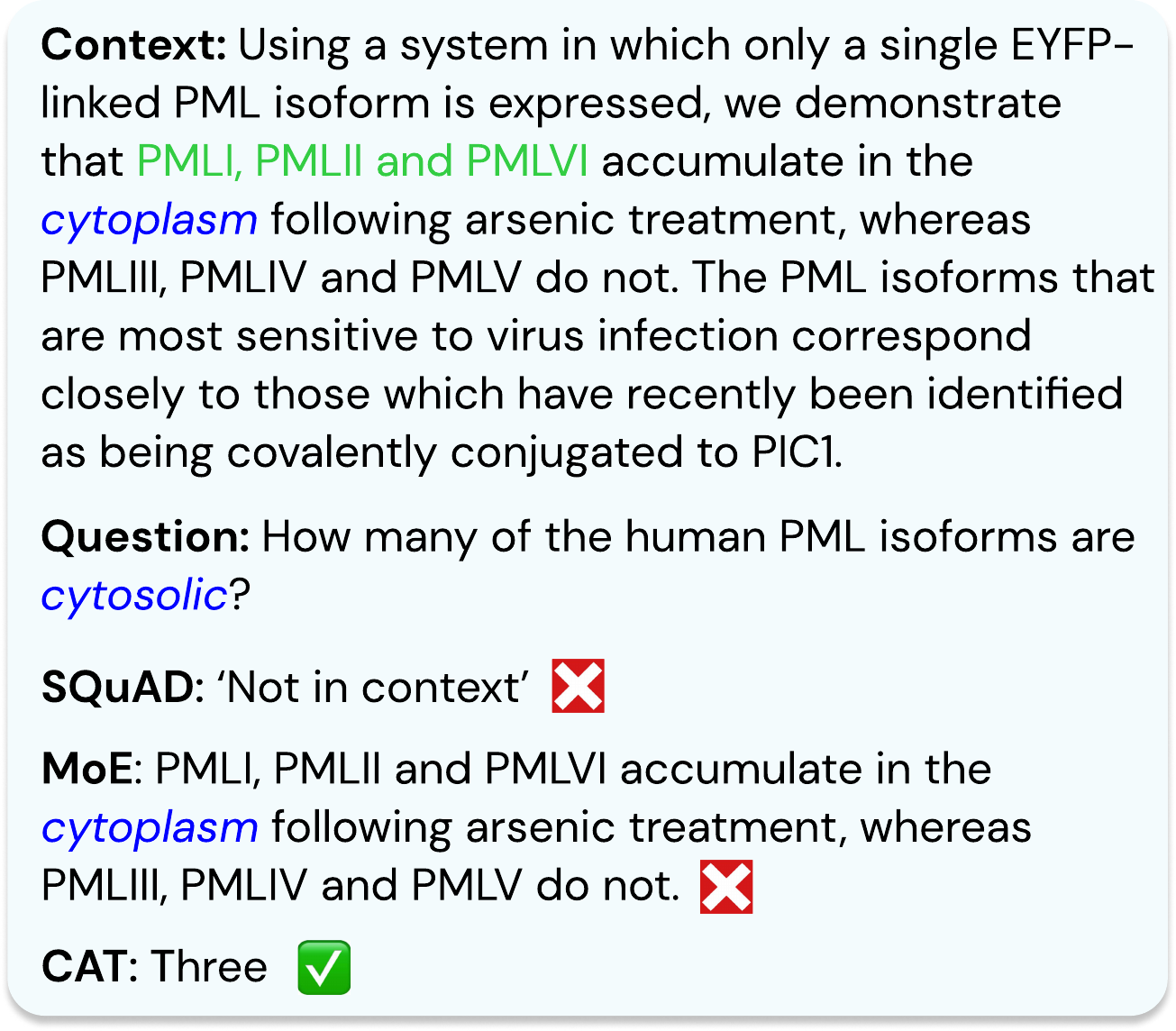}
    \caption{\small Answers generated by CAT, MoE, and model trained on SQuAD when asking a BioASQ question.}
    \label{fig:examples_rc}
\end{subfigure}

\caption{Quantitative and qualitative results on reading comprehension task. }\label{fig:reading_comprehension}

\vspace{-.5cm}

\end{figure}

\paragraph{Results.} \autoref{fig:reading_comprehension} reports the pairwise win fractions and ELO ratings obtained by the different methods on BioASQ. Again, we observe that CAT is the best method both in terms of pairwise win fraction and ELO ratings. Evidently, the question-answering skill is more useful than having domain knowledge as SQuAD fairs $6\times$ better compared to textbook. From \autoref{fig:examples_rc} we see the lack of biomedical knowledge hurting SQuAD; how CAT gives well-formed answers compared to MoE which is indeed able to understand the context but just copies relevant text. The other methods are worse and the merging methods beat DATA-MIX. As in the previous experiment, we use continual learning for the DATA-MIX experiment.

\subsection{LoRA merging improves robustness to prompt format changes}\label{sec:prompt_robustness}

\begin{wrapfigure}[14]{r}{0.45\textwidth}
\vspace{-.5cm}
\includegraphics[width=0.45\textwidth]{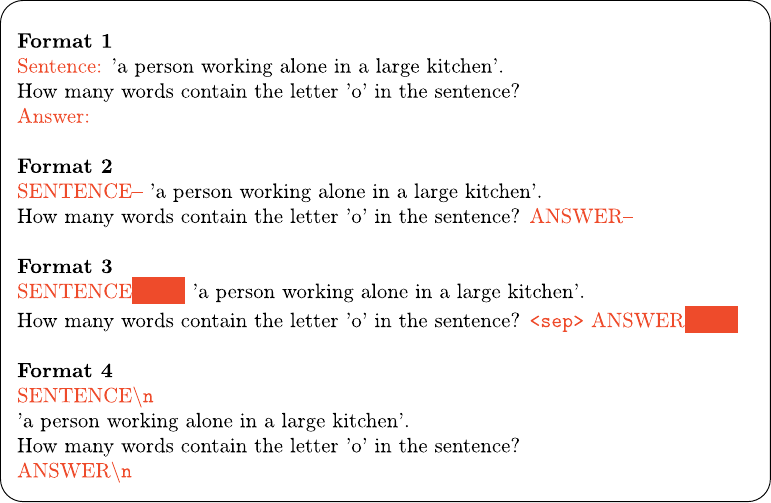}
\caption{\small Different prompt formats -- varying  \textit{descriptors}, \textit{separators}, and \textit{spaces} for the same task example.}\label{fig:prompts_examples}
\end{wrapfigure}

\paragraph{Problem.}  Recently, a few works \cite{mizrahi2024state,sclar2024quantifying} reported the sensitivity of language models to prompt formatting. These works report performance differences of up
to 76 accuracy points due to subtle changes
in prompt formatting when using LLaMA-2-13B \cite{touvron2023llama}. This sensitivity remains when increasing model size, the number of few-shot examples, or performing instruction tuning. We investigate whether model or data mixing is a solution to this problem. Therefore, we set the first skill to be the dataset with format $i$ and the second skill to be the dataset with format $j$ and evaluate the merging methods on a dataset with format $k$, where $i,j,k\in\{1,\dots,10\}$ and $i\neq j\neq k.$


\paragraph{Evaluation setup.} We choose \texttt{task\_131} of counting the number of words having a particular letter in the given sentence from Super-NaturalInstructions \cite{supernaturalinstructions}. Following the grammar over \textit{descriptors}, \textit{separators}, and \textit{spaces} defined in \cite{sclar2024quantifying}, we sample 7 prompt formats (see \autoref{fig:prompts_examples} for some examples of format variations -- as simple as spacing, casing). The \textit{descriptors} refer to the format of \textit{\{Sentence, Answer\}}, \textit{separators} refer to the separator between the \textit{descriptor} and the example, and \textit{space} referees to the dividing character between the \textit{descriptors}. We train on 6000 examples and test on 417.

\paragraph{Results.} \autoref{fig:prompt_robustness} reports the performance of each model when evaluating on 4 out-of-distribution formats (Formats 3, 7, 8 and 10). Figures \ref{fig:prompt_format14}, \ref{fig:prompt_format12} and \ref{fig:prompt_format24} respectively display the single and merged models when trained on Formats 1\&4, Formats 1\&2 and Formats 2\&4. A well-performing merged model should perform at least as well as the individual models i.e.\ the models trained on a single format. 
Remarkably, data mixing performs poorly, even worse than the individual models. Regarding the merged models, the results are variable. When finetuning on Formats 1\&4 (\autoref{fig:prompt_format14}), the performance of CAT does not vary across target formats and remains high. TIES performance is also high contrary to DARE. On Formats 1\&2 (\autoref{fig:prompt_format12}), TIES is the best model since it performs as the single model when evaluated on Formats 3 and 10. CAT and DARE perform worse and only obtain a decent performance when evaluated on Format 3. Lastly, on Formats 2 \& 4, merging fails since all the models perform poorly.
Thus, we show that model merging is an approach to attaining robustness to prompt formatting changes.

\subsection{Ablations}
\label{sec:ablation}

\begin{wrapfigure}[12]{r}{0.45\textwidth}
\vspace{-1.5cm}
\includegraphics[width=0.45\textwidth]{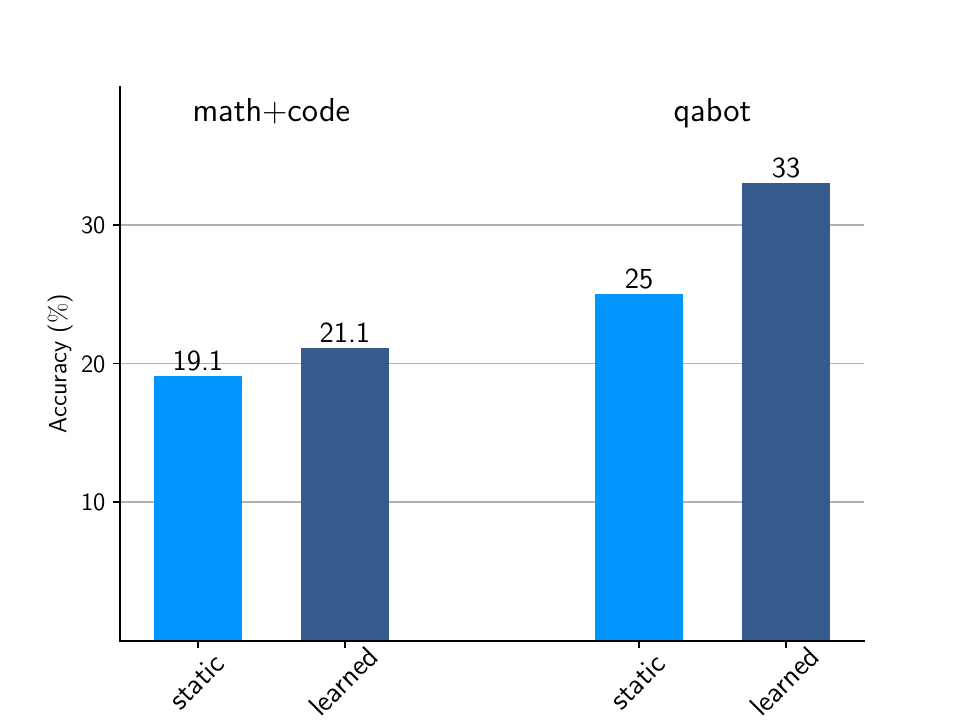}
\caption{\small Performance of the learned vs static approach in CAT.  }\label{fig:ablation}
\end{wrapfigure}

\paragraph{Learning the weights of CAT.}
We analyze the impact of learning the weights to be assigned to each skill. As discussed in \autoref{s:exp_details}, we learn the merging coefficients. In this section, we demonstrate that this learning entails a significant performance improvement. \autoref{fig:ablation} shows how that ``learned'' CAT beats ``static'' CAT in the GSM-Hard experiment and the QAbot experiment (on the game manual). In this case, we simply average -- set $\alpha_1^l,\alpha_2^l=0.5$.

\section{Limitations \& Future Directions}\label{sec:limitations}

Our work is a first step towards using model merging for solving skill composition tasks. Our study is limited to settings where the number of needed skills $k=2$. Extension to a higher number of skills is challenging for two reasons:  (1) Identifying practical settings that need a large number of skills that can be clearly identified is difficult. (2) Even if we find such settings, it remains unclear whether LoRA merging is effective. To test a hypothetical situation, we consider the prompt robustness experiment from \autoref{sec:prompt_robustness} and report the performance of the different baselines when trained on 3 different prompt formats. This time, we observe that all the LoRA merging baselines perform worse than DATA-MIX (refer \autoref{appx:prompt_robust} \autoref{fig:limitations}). This suggests that when the number of skills is large, DATA-MIX is still the best strategy for skill composition. Improving current model merging schemes is an important future direction for getting an efficient composition of many skills.

\section{Conclusion}
We conclude that when obtaining training data is challenging, decomposing a task into its underlying skills and model merging is a promising approach. We demonstrate several practical use cases that can be treated as  binary skill composition problems.  We show that methods based on the concatenation of LoRAs achieve the best skill composition and outperform the popular data mixing method. Another highlight of our analysis is the super-linear modularity results from \autoref{sec:math_code} which shows that the model is effectively able to compose the math and code skills to solve new problems. As mentioned in \autoref{sec:limitations}, an exciting future direction is to design merging methods that would enable to composition of more than two skills. This would give an interesting alternative to the current paradigm where we train large-scale models on curated datasets; instead we could merge specialized LoRAs to obtain an equivalently good model.

\bibliography{references}
\bibliographystyle{abbrvnat}


\appendix

\section{Review on LoRA}\label{app:lora}

All the methods we study for solving composition skill problems are based on LoRA \cite{hu2021lora}, which is defined as follows. During finetuning, the update of the weights are constrained to be a low-rank decomposition i.e.\  the update is $W_0+\Delta W$, where $\Delta W= BA^{\top}$, for $W_0$ pre-trained weights,  $B\in\mathbb{R}^{d\times r}$ and $A\in\mathbb{R}^{d\times r}$ trainable parameters, and $r\ll d.$ 

LoRA presents several advantages compared to standard finetuning. First, it is more parameter-efficient i.e., it uses a lower number of trainable parameters and has a lower memory usage i.e.,  fewer parameters need to be stored and processed which lowers the memory footprint. More importantly, it is more modular i.e.\  LoRA's method of isolating additional parameters makes it easier to manage adaptations and switch between different fine-tuned tasks. It is possible to load and apply different sets of low-rank adaptations without needing to retrain the entire model from scratch for each new task. For these reasons, we focus on LoRA-based methods to solve skill composition problems. We detail them in the next section.

\section{Additional experimental details}
\label{appx:additional_exp_details}

\subsection{Training details \& computing resources.}
\label{appx:compute}
We run our experiments on 
the following GPUs depending on their availability on our compute cluster: NVIDIA RTX A6000, NVIDIA RTX A5000, and NVIDIA A100.
Mainly, for the most extensive large-scale fine-tuning of five textbooks for reading comprehension, we train the models in a distributed multi-GPU environment using DeepSpeed on 2 A100 GPUs. This takes about 2 days. 

For reading comprehension, we use a sequence length of 2048, while for question-answer we found that using smaller lengths in \{20,100\} worked better at memorizing minute details.

In all data and LoRA merging methods, we do 1-shot prompting to ensure the output is of consistent format. 

\subsection{Datasets introduced.} 
\label{appx:qa_data}
For the QABot task, we create two new datasets consisting of 95 examples (45 from the biology textbook \url{} chapter 16 and 50 from the rules of \textit{Dungeon \& Dragons} manual). We extract text contents from these to prepare the training corpus. This amounts to 8169 tokens in the case of biology and 25427 tokens for the game. 
To obtain question-answer pairs, we provide page-level content in the context and use the Question-Answer generation prompt shown in \autoref{fig:gpt4_prompts} to prompt GPT-4 \cite{achiam2023gpt}.
This is followed by a manual inspection to ensure valid and appropriate questions.  

\section{Additional experiments}
\label{appx:additional_exp}

\subsection{Math-word problems.}
\label{appx:math-word}
Since this setting requires training over smaller scale data than reading comprehension and QABot, we conducted an evaluation to assess the robustness of the reported accuracy metric of the CAT method. We obtain $21.63 \pm 1.07$ when testing with 3 different seeds. which is still greater than DATA-MIX (18.8).

\subsection{QABot}
\label{appx:chatbot}

\begin{wrapfigure}[11]{r}{0.45\textwidth}
\vspace{-1.7cm}
\includegraphics[width=0.45\textwidth]{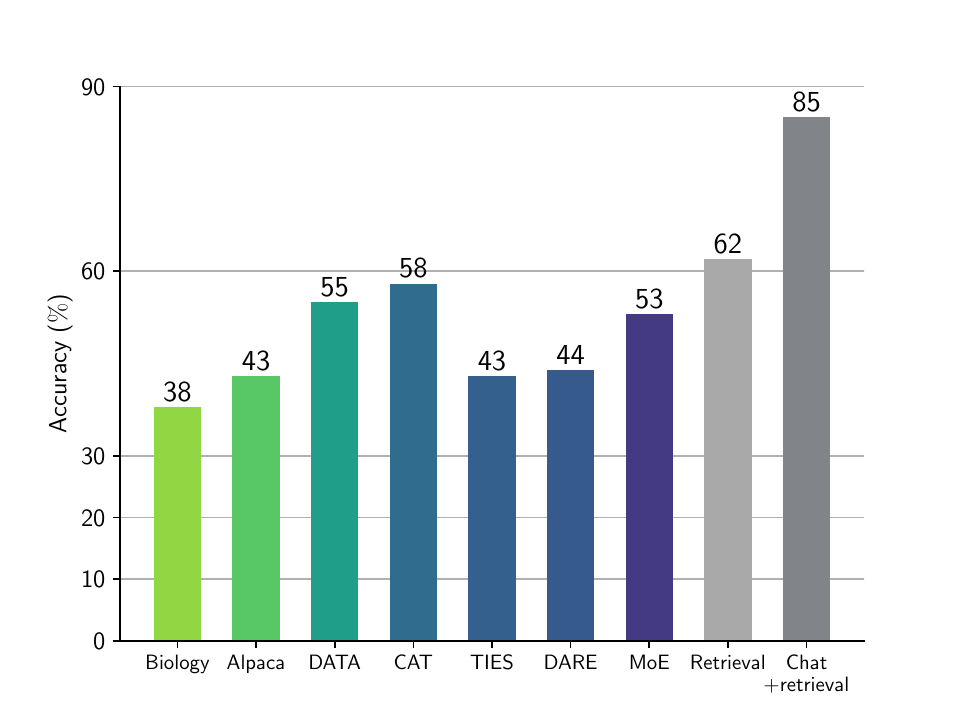}
\caption{\small QABot on biology chapter.}\label{fig:chatbot_bio}
\end{wrapfigure}

For judging, we use accuracy as the metric using the QA judge prompt in \autoref{fig:gpt4_prompts} .

\textbf{Biology textbook QA.}
In the simpler setting of a chapter from a university biology textbook, we see that the base Llama model that has been instruction fine-tuned on Alpaca is able to obtain 43\%, which is enhanced by DATA-MIX which injects domain knowledge to 54\%. 

\subsection{Reading comprehension}
\label{appx:reading_comp}

\begin{wraptable}[10]{r}{0.45\textwidth}
\vspace{-.9cm}
\centering
\resizebox{.3\columnwidth}{!}{%
\begin{tabular}{ll}
\toprule
\bfseries Model & \bfseries F1 score \\
\midrule
  textbook (5-shot)        &     0.041     \\
   SQuAD       &     0.047     \\
   TIES       &      0.04    \\
    DARE      &    0.037      \\
 MoE      &   0.037           \\
 DATA & 0.044 \\
 CAT      &     0.028     \\
\bottomrule
\end{tabular}
}
\caption{\small F1 scores on BioASQ.}\label{tab:bioasq_f1}
\end{wraptable} 


The corpus of textbooks used to impart biomedical knowledge contains 1417501 tokens.

As discussed in \autoref{sec:reading_comprehension}, since the gold reference answers in BioASQ are quite descriptive unlike the simpler/concise answers for the QABot datasets, the naive F1 based scoring is unable to reflect the true performance of models \autoref{tab:bioasq_f1}. we observed that when asked to score a model individually, scores from GPT-4 do not capture the fine-grained details or consider relative generations from other models. Hence, we resort to pairwise scoring similar to LMSys. Using this scheme gives us more reliable scores. For ELO computation, we start with an initial rating of 200, base 10, scale 400, and $K$-factor $4$. We bootstrap the ELO ratings $5000$ times to ensure stable results.

\subsection{Prompt robustness}
\label{appx:prompt_robust}

\begin{figure}[h]
\begin{subfigure}{0.30\textwidth}
\hspace*{-.4cm}\includegraphics[width=1.15\linewidth]{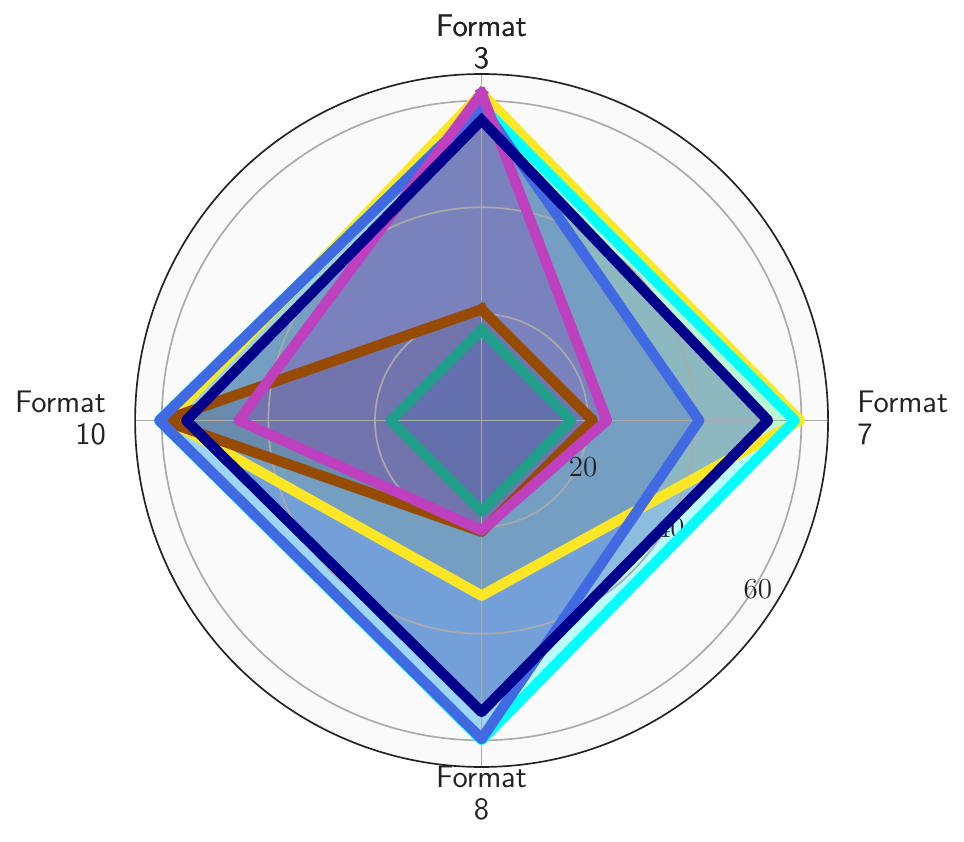}
\caption{Format 1\&4}\label{fig:prompt_format14}
\end{subfigure}
\begin{subfigure}{0.30\textwidth}
\hspace*{-.2cm}\includegraphics[width=1.15\linewidth]{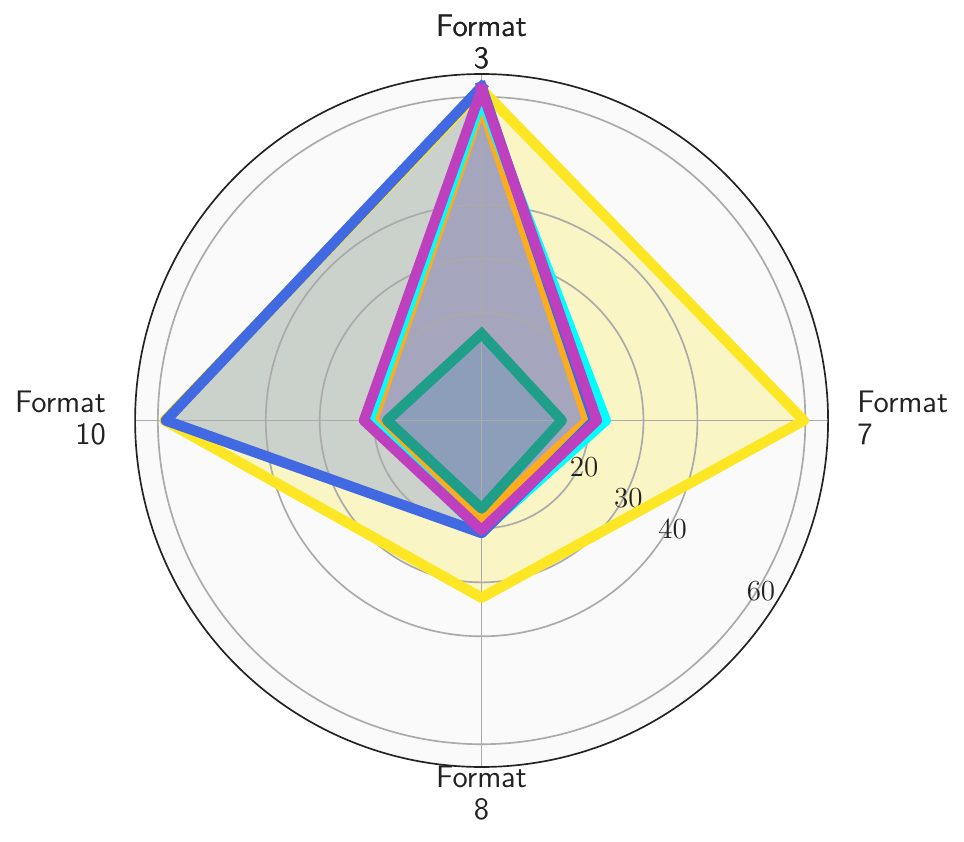}
\caption{Format 1\&2}\label{fig:prompt_format12}
\end{subfigure}
\begin{subfigure}{0.30\textwidth}
\hspace*{-.0cm}\includegraphics[width=1.15\linewidth]{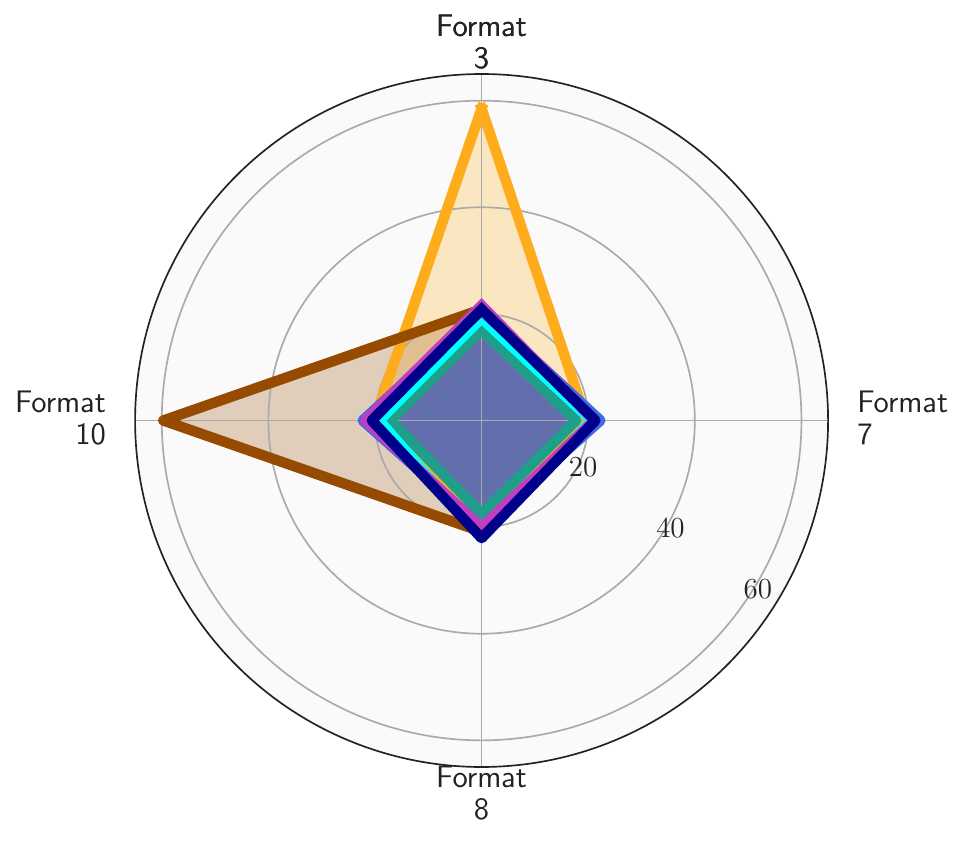}
\caption{Format 2\&4}\label{fig:prompt_format24}
\end{subfigure}
\small 
{\hspace*{3cm}\mbox{Single LoRA:\hspace*{.5cm}\cblock{253}{231}{37} \hspace{0.5mm}Format 1\hspace{1mm} \cblock{255}{87}{51}\hspace{0.5mm}Format 2\hspace{1mm} \cblock{150} {75}{0}\hspace{1mm}Format 4}}\\ 
{\hspace*{3cm}\mbox{Merging:\hspace*{1.1cm}\cblock{31}{158}{137} \hspace{0.5mm}DATA-MIX\hspace{1mm} \cblock{0}{255}{255} \hspace{0.5mm}CAT\hspace{1mm} \cblock{65}{105}{225}\hspace{1mm}TIES\hspace{1mm} \cblock{191}{64}{191}\hspace{1mm}DARE\hspace{1mm}\hspace{1mm}\cblock{0}{0}{139}\hspace{0.5mm} MoE}}
\caption{\small Performance of single LoRAs and merged models trained on format pairs mentioned and tested on different formats. }\label{fig:prompt_robustness}

\vspace*{-.5cm}

\end{figure}

\paragraph{Prompt format robustness. } In \autoref{fig:prompt_robustness}, we see different merging methods working well for different format pairs trained. While we do not see a clear strategy that would guarantee robustness, it indicates that merging methods are capable of attaining robustness. Achieving this is a very desirable phenomenon as this would eliminate the need to prompt engineer by trying diverse formatting choices.

\begin{figure}[h!]
\centering
\includegraphics[width=0.35\textwidth]{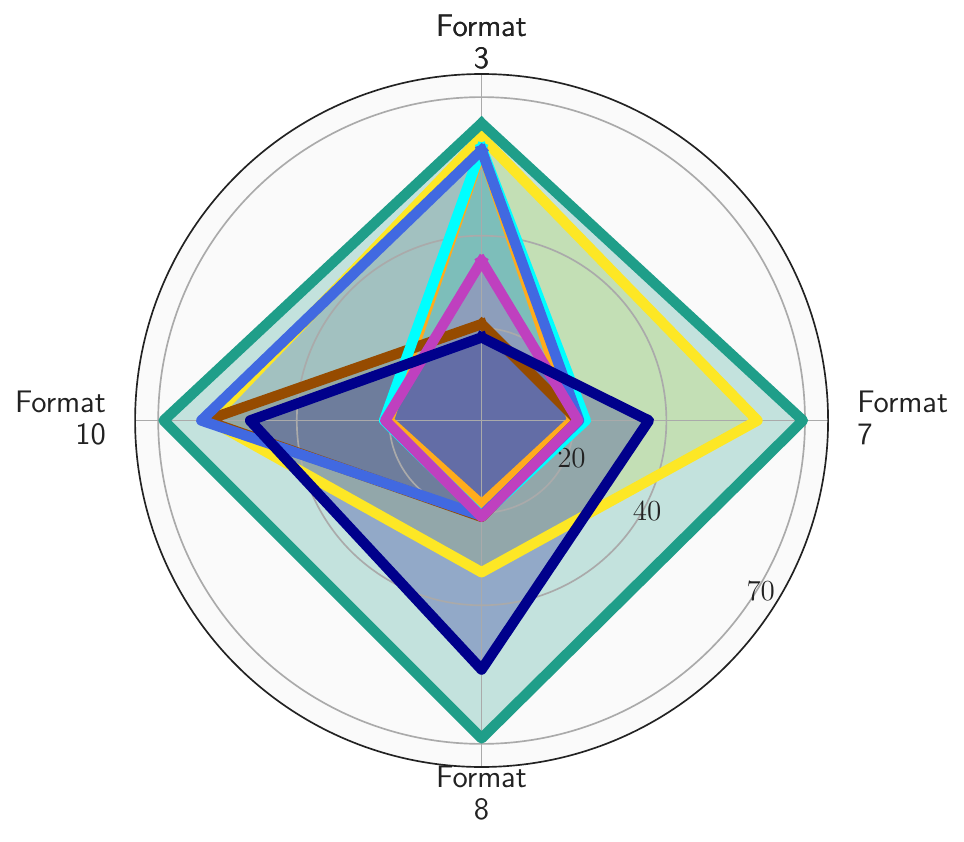}
\caption{\small Testing prompt robustness when training on 3 formats.}\label{fig:limitations}
\end{figure}

\paragraph{Merging on 3 formats.}
\autoref{fig:limitations} shows the performance of various methods when 3 formats are combined. As we can see, DATA-MIX clearly beats the other methods across all formats tested. 

\begin{figure}
 \begin{minipage}[b]{0.27\textwidth}
        \centering
\begin{tcolorbox}[
    width=\textwidth,
    fontupper=\ttfamily,
    left=0.5mm,
    right=0.5mm,
    top=0.5mm,
    bottom=0.5mm,
    title=Question-answer generation prompt,
    center title,
    fonttitle=\tiny\ttfamily]
\tiny
<QUESTION PROMPT>\\
You are an expert in generating questions. Based on the given Context, please generate a question whose answer lies in the Context. The question should be to the point and must entail a definite, concise answer in the Context. Focus on specific points and details such that only someone who understands the Context well can answer.\\

Context: \{context\}\\

<ANSWER PROMPT>\\
You are an expert at reading comprehension. Given the Context, please respond to the Question based on the information in the Context. The answer should at max be one sentence long. Output only the exact answer.
\end{tcolorbox}
\end{minipage}
\hfill
    \begin{minipage}[b]{0.36\textwidth}
        \centering
        \begin{tcolorbox}[
    width=\textwidth,
    fontupper=\ttfamily,
    left=0.5mm,
    right=0.5mm,
    top=0.5mm,
    bottom=0.5mm,
    title=Reading comprehension judge prompt,
    center title,
    fonttitle=\tiny\ttfamily]
\tiny
Please act as an impartial judge and evaluate the quality of the answers.\\

You will receive five items: Context, Question, Gold Answer, Choice 1, and Choice 2. Your task is to assess which among Choice 1 and Choice 2 in terms of CORRECTNESS. You MUST read and understand the Context and assess the Choice answers with respect to the information in the Context. To help you with evaluation, we provide the Gold Answer. The Gold Answer has been checked by experts and is 100\% correct. Use it as a reference for spotting CORRECTNESS errors.\\

You should give a score from 0 to 3 for CORRECTNESS to Choice 1 and Choice 2. Half points are allowed. Then you must say which choice was better.\\

Here is the Context:\\
\{context\}\\
Here is the Question:\\
\{question\}\\
Here is the Gold Answer:\\
\{answer\}\\
Here is Choice 1:\\
\{choice1\}\\
Here is Choice 2:\\
\{choice2\}\\

Please present your scores as follows:\\
Choice 1: x/3\\
Choice 2: x/3\\
Choice \_ is better
\end{tcolorbox}
\end{minipage}
\hfill
    \begin{minipage}[b]{0.34\textwidth}
        \centering
\begin{tcolorbox}[
    width=\textwidth,
    fontupper=\ttfamily,
    left=0.5mm,
    right=0.5mm,
    top=0.5mm,
    bottom=0.5mm,
    title=Question-answer judge prompt,
    center title,
    fonttitle=\tiny\ttfamily]
\tiny
Please act as an impartial judge and evaluate the quality of the answers.\\

You will receive four items: Context, Question, Gold Answer, and System Answer. Your task is to assess the System Answer in terms of CORRECTNESS. You MUST read and understand the Context and assess the System Answer with respect to the information in the Context. To help you with evaluation, we provide the Gold Answer. The Gold Answer is an approximately correct answer. Use it as a reference for spotting CORRECTNESS errors but you should judge based on the provided Context.\\

You should give a score from 0 to 2 for CORRECTNESS to the System Answer.\\

Here is the Context:\\
\{context\}\\
Here is the Question:\\
\{question\}\\
Here is the Gold Answer:\\
\{gold\_answer\}\\
Here is the System Answer:\\
\{system\_answe\}\\
Here is Choice 2:\\
\{choice2\}\\

Please present your scores as follows:\\
Score: x
\end{tcolorbox}
\end{minipage}
\caption{Prompts used to generate questions/judge answers using GPT-4.}
\label{fig:gpt4_prompts}
\end{figure}




\end{document}